%% file: main.tex
\documentclass[sigconf]{acmart}
\copyrightyear{2022}
\acmYear{2022}
\setcopyright{acmlicensed}
\acmConference[WoRMA '22]{Proceedings of the 1st Workshop on Robust Malware Analysis }{May 30, 2022}{Nagasaki, Japan}
\acmBooktitle{Proceedings of the 1st Workshop on Robust Malware Analysis (WoRMA '22), May 30, 2022, Nagasaki, Japan}
\acmPrice{15.00}
\acmDOI{10.1145/3494110.3528242}
\acmISBN{978-1-4503-9179-5/22/05}

\usepackage{mathtools}
\newcommand{\defeq}{\vcentcolon=}
\usepackage{pdfpages}
\usepackage{algorithm}
\usepackage{algorithmic}
\usepackage[color=green]{todonotes}
\usepackage{multirow}
\usepackage{caption}
\usepackage[utf8]{inputenc}
\usepackage{booktabs}
\usepackage{balance}
\usepackage{pgfplots}
\usepackage{graphicx}
\usepackage{caption}
\usepackage{subcaption}
\usepackage{pgfplots, pgfplotstable}
\pgfplotsset{compat=1.14}
\usetikzlibrary{pgfplots.groupplots,shapes,decorations}

\newlength\mylinewidth
\newlength\figureheight
\newlength\figurewidth
\newlength\smallfigureheight
\newlength\smallfigurewidth
\newlength\largefigureheight
\newlength\largefigurewidth
\newcommand{\newtext}[1]{\textcolor{black}{#1}}
\newcommand{\oldtext}[1]{\ignorespaces}

\makeatletter
\renewcommand\paragraph{\@startsection{paragraph}{4}{0mm} % name, level, indent 
{0.4\baselineskip} % beforeskip: how much space you want before \paragraph 
{-0.6em} % afterskip: how much space you want after \paragraph 
{\normalfont\bfseries%   the font family, etc. you want for \paragraph 
}}% 

\usepackage{geometry}
\begin{document}
\fancyhead{}

\title{Transformers for End-to-End InfoSec Tasks: A Feasibility Study}

%\author{Anonymous}
\author{Ethan M. Rudd}
\affiliation{%
  \institution{Mandiant Inc.
  }
  \city{Reston, VA}
  \country{USA}
  }
\email{ethan.rudd@mandiant.com}

\author{Mohammad Saidur Rahman}
\authornote{Work done while at Mandiant Inc.}
\affiliation{%
  \institution{Rochester Institute of Technology
  }
  \city{Rochester, NY}
  \country{USA}
  }
\email{saidur.rahman@mail.rit.edu}

\author{Philip Tully}
\affiliation{%
  \institution{Mandiant Inc.
  }
  \city{Reston, VA}
  \country{USA}
  }
\email{philip.tully@mandiant.com}

% in the abstract
\input{0_abstract}



% make the title area
\maketitle

\input{1_Introduction}

\input{2_Background}

\input{3_Method}
\input{3.1_Byte_Approach}

\input{4_Evaluation}
\input{4.1_Byte_Evaluation}

\input{5_Discussion}

\input{6_Conclusion}
\input{7_Acknowledgement}

\balance

\bibliographystyle{ACM-Reference-Format}
\bibliography{transformers}

\end{document}

%% file: 0_abstract.tex
\begin{abstract}
    Training a machine learning (ML) model from raw information security (InfoSec) data involves utilizing distinct data types and input formats that require unique considerations compared to more conventional applications of ML like natural language processing (NLP) and computer vision (CV). In this paper, we assess the viability of transformer models in end-to-end InfoSec settings, in which no intermediate feature representations or processing steps occur outside the model. We implement transformer models for two distinct InfoSec data formats – specifically URLs and PE files – in a novel end-to-end approach, and explore a variety of architectural designs, training regimes, and experimental settings to determine the ingredients necessary for performant detection models. 
    
    We show that in contrast to conventional transformers trained on more standard NLP–related tasks, our URL transformer model requires a different training approach to reach high performance levels.  Specifically, we show that 1) pre-training on a massive corpus of unlabeled URL data for an auto-regressive task does not readily transfer to binary classification of malicious or benign URLs, but 2) that using an auxiliary auto-regressive loss improves performance when training from scratch. We introduce a method for mixed objective optimization, which dynamically balances contributions from both loss terms so that neither one of them dominates. We show that this method yields quantitative evaluation metrics comparable to that of several top-performing benchmark classifiers. 
    
    Unlike URLs, binary executables contain longer and more distributed sequences of information-rich bytes. To accommodate such lengthy byte sequences, we introduce additional context length into the transformer by providing its self-attention layers with an adaptive span similar to Sukhbaatar et al. We demonstrate that this approach performs comparably to well-established malware detection models on benchmark PE file datasets, but also point out the need for further exploration into model improvements in scalability and compute efficiency.
  
\end{abstract}

%% file: 1_Introduction.tex
\section{Introduction}

The abundance of labeled data sources has been a major driver in the success of ML in various applications such as image recognition~\cite{krizhevsky2012imagenet}, audio recognition~\cite{hinton2012deep}, natural langugage processing (NLP) \cite{devlin2018bert}, and InfoSec~\cite{raff2018malware, pascanu2015malware, li2019mad, sirinam2019triplet, rahman2020mockingbird}. For applications which involve image, natural language text, or audio data, a recent dominating trend has been to feed raw data with increasingly minimal preprocessing directly to an ML model in favor of fitting a model on hand-engineered features. For many applications, this approach of learning from raw data works well. Applications of ML in InfoSec, however, require  dealing with unique data formats which must be approached in different ways, and this is one of the reasons why InfoSec applications of ML still rely heavily on relatively simple models fit on hand-engineered features.

This may be in part because properly learning features based on raw data is less trivial for some of the data formats inherent to security problems, for example, in  \cite{anderson2018ember}, a convolutional neural network (CNN) which operates on raw PE files and is closely analogous to CV or NLP CNN \cite{raff2018malware} under-performs a gradient boosted decision tree model trained with default parameters on hand-engineered features. 

From a scalability perspective, hand-engineered features for InfoSec tasks are non-trivial because creating them often requires expensive and time-consuming interaction with subject matter experts (SMEs). Moreover, ongoing attacker innovations mean that feature extractors may need to be continuously updated in order to keep pace. Models that learn features directly from raw data, on the other hand, require no such updates to feature extraction processors and do not require as much SME interaction. Thus, there is substantial motivation for research into more effective ML models for InfoSec tasks that operate on raw data.

With respect to other domains where ML models operating on raw data have advanced the state of the art, NLP is arguably most analogous to InfoSec. Some InfoSec tasks can even be framed as NLP problems, including email spam detection~\cite{dada2019machine}, website content categorization~\cite{rao2019detection}, social media phishing~
\cite{seymour2016weaponizing}, and certain areas of data leak prevention (DLP)~\cite{alzhrani2016automated,alzhrani2016automated2,alzhrani2017automated}. When framing InfoSec tasks in NLP terms, the raw data does not necessarily conform to typical prosaic sentence and paragraph structures. Some InfoSec tasks, e.g., source code attribution \cite{alsulami2017source}, contain little to no ``natural language" yet may possess their own linguistic structure. InfoSec problems also deviate from NLP problems in terms of how their classifiers are typically trained. InfoSec classifiers in the industry are typically fit on millions to hundreds of millions of weakly labeled samples, where supervisory signals are e.g., aggregated over threat feeds or externally derived vendor scores. NLP classifiers, particularly those which rely on transformers, typically use large datasets for self-supervised pre-training of a base architecture, but relatively small datasets for fine tuning final layers for a given classification task. We explore the ramifications of this within this paper. 

InfoSec data also more commonly contains long-ranging sequential dependencies in which one or more tokens far-removed from a given token may strongly influence the probability of that token. Given the close analogy to NLP, and because transformer architectures have revolutionized performance in the NLP domain, it seems fruitful to explore the feasibility of applying transformers to InfoSec tasks, especially since they address the long range dependency issue using an attention mechanism. 

However, InfoSec is a broad field, and sequential dependency ranges vary dramatically based on the data type.  
In NLP literature, “long-range” dependencies are typically 
considered to span tens to tens of thousands of tokens – often backwards -- within a sequence. Similar ranges may apply for certain types of InfoSec tasks, but not for others. For example, binary classification, where function calls and declarations can be almost arbitrarily separated may involve relevant dependencies spanning forward or backward millions or even billions of bytes.

While research has been conducted on scaling transformer context window sizes, additional research is required to develop transformer architectures that work well for a broad array of InfoSec tasks, specifically those with their own data formats that deviate from conventional applications of NLP. In this paper, we conduct a feasibility study using two different data formats: URLs and PE files. Other works \cite{raff2018malware,coull2019activation,saxe2017expose} have applied deep learning based on raw data from these formats and explored the learnt representations. Applying transformers towards these tasks is in many ways a natural extension of this line of research. For each data format we perform a malicious/benign detection task. For URLs, we aim to detect those linking to malicious content, while for PEs we aim to detect files containing malware. We chose to focus on these two data formats because they are not natural language but have distinctly different characteristics under which we can explore the viability of transformer models: URLs are relatively short sequences of characters, can fit nicely into a canonical transformer’s context window, and allow for relatively efficient training on a GPU cluster. This allows us to explore the effects of different training strategies and loss functions under feasible iteration times. In contrast, PE files 
span long sequences of bytes, with large quantities of content that potentially provides little to no malicious or benign indications (e.g., padding, images, or encrypted payloads). 
Thus, modifications for longer sequences must be made to the canonical transformer to feasibly operate on raw bytes (e.g., hidden state caching~\cite{dai2019transformer}, sparse attention patterns~\cite{child2019generating}, locality sensitive hashing~\cite{kitaev2020reformer}, etc.).

The contributions of our feasibility study are as follows:

\begin{itemize}
    
\item A performance comparison between our novel transformer approaches and other more conventional approaches to URL classification. Using a transformer model, we are able to achieve performance on par with our top benchmark models.
    
\item A performance comparison of multiple URL transformer training regimes, including with and without auto-regressive pre-training. We demonstrate that in contrast to NLP applications which rely heavily on
self-supervised~
%unsupervised 
pre-training, this strategy does not readily improve performance on the URL classification task.
    
\item A novel balanced mixed objective transformer loss function which balances a classification objective with an auxiliary auto-regressive objective during training. This is designed to encourage the transformer to fuse both sequential context with class information in its hidden states.

\item A performance comparison between our own modification of an adaptive attention span transformer~\cite{sukhbaatar2019adaptive} and other byte-based and feature-based models on truncated PE files from the EMBER dataset. This includes featurizing the EMBER PE files after truncation to offer a fair comparison, which has typically not been performed in the literature when comparing raw byte models to featurized models.

\item An exploration of various ways to encode bytes with little to no information loss for raw-byte models, including transformers. We investigate whether or not these encodings may be able to improve performance by extending effective context window size at the cost of increased vocabulary size. 

\end{itemize}

%% file: 2_Background.tex
\section{Background}
\label{sec:background}

\begin{figure*}[ht]
\centering
\subfloat[Next Character Prediction]{\includegraphics[width=0.32\textwidth]{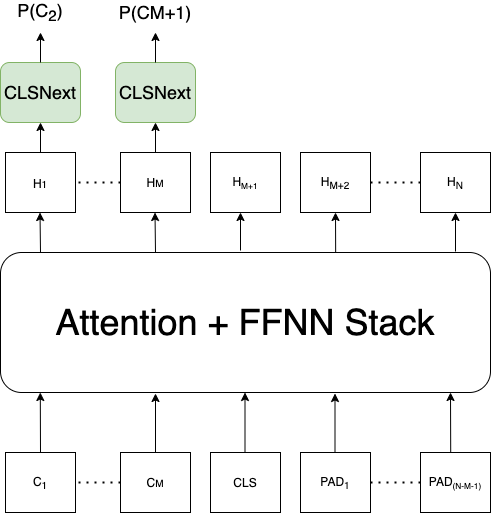}\label{fig:next}}\hfill
\subfloat[Classification Only]{\includegraphics[width=0.32\textwidth]{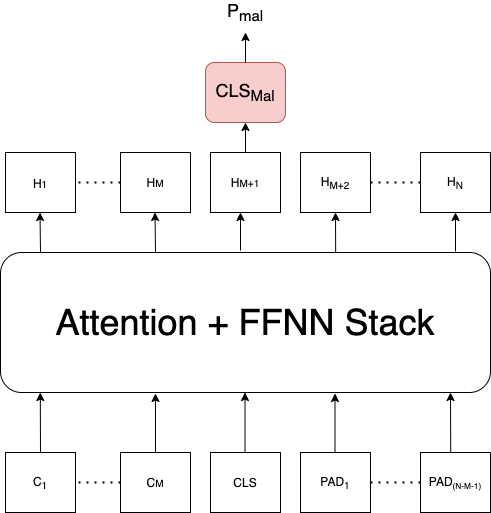}\label{fig:cls}}\hfill
\subfloat[Classification with Auxiliary Loss]{\includegraphics[width=0.32\textwidth]{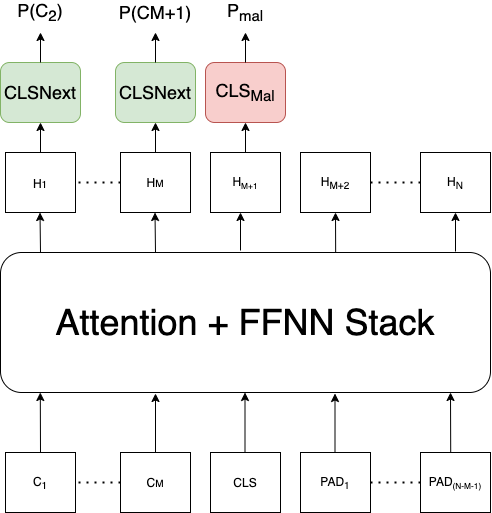}\label{fig:mixed}}
\caption{For a given context window size, transformers utilize padding/masking operations to perform inference over different sequence lengths. Let $M$ be the length of the sequence and $N$ be the size of the context window, where $M \leq N-1$. Let $C_1$ through $C_M$ be the embeddings corresponding to the sequence elements, in this case, characters within a URL. Let CLS be the embedding corresponding to the CLS token (position $M+1$), and let $PAD_1$ through $PAD_{N-M-1}$ be embeddings of padding tokens. Let $H_1$ through $H_N$ be the final hidden states derived from the transformer. A standard \textit{next character prediction} task is shown in  \protect\subref{fig:next}, where the hidden states are each passed through a classifier $CLS_{Next}$, and the softmax outputs correspond to probabilities of the subsequent tokens. \protect\subref{fig:cls} We can also use the output of $H_{M+1}$, passed through a binary classifier to derive a probability that the input sequence is malicious or benign. \protect\subref{fig:mixed} During training, we can use all token outputs to fit a malicious/benign predictor and a next-character prediction loss simultaneously. Note that hidden states corresponding to padding are ignored. For our implementation, left-to-right masking operations prevent prior hidden states from attending to any padding-related sequence information.}
\label{fig:transfomer}
\end{figure*}

Transformers, introduced by Vaswani et al.~\cite{vaswani2017attention} have revolutionized NLP and a variety of other discrete sequence modeling domains. This is partly due to their ability to directly incorporate long-term dependencies in a sequence and partly because they are easily trained in parallel. Transformers were originally formulated for sequence transduction tasks, e.g., neural machine translation (NMT)~\cite{bahdanau2014neural}, wherein a source sequence is encoded via an encoder stack of attention and feed-forward layers, then decoded via a similar decoder stack, using the input sequence as a source context for decoding to the target sequence. %E.g., in an English to French translation model, the encoding of the English sentence would serve as context for the French translated sentence; during translation, the context encoding along with hidden states from the translated portion of the sentence are used to arrive at the next token value.
Later models, which constitute massively pre-trained representations, meant for fine-tuning for non-transductive tasks abandon the encoder/decoder setup entirely, opting in favor of either an encoder or a decoder. Depending on the problem, the flow of information from the input sequence may be either bidirectional, e.g., \cite{devlin2018bert}, or left to right (L-R) \cite{radford2018improving}. 

While transformers can be implemented in a variety of ways, a typical implementation of a transformer attention stack consists of the following components:

\begin{enumerate}
%\item Embedding: The input, in our case a URL, is projected into an embedding space.
    \item Embedding: The input is projected into an embedding space. 
    \item Positional Encoding: This is a signal added to each of the embedding vectors to imbue each embedding with a positional order in the sequence (which would otherwise not be tracked by the attention mechanism). This can optionally be added to hidden states within the transformer, but we do not do this in our implementation.
    \item Multi-Headed Attention Layers: These apply the attention mechanism in parallel in a redundant fashion across $h$ heads. The results are concatenated and fed to a feed-forward layer.
    \item Feed-Forward Layers: These act on each of the hidden states produced by multi-headed attention.
    \item Residual Connections: These add and norm operations occur between inputs to each stack of attention/FFNN layers and their outputs. This allows information to percolate up the transformer layers, bypassing particular attention/FFNN blocks when appropriate.
    \item Masks: Because transformers operate in parallel, padding is typically added to inputs shorter than the context window of the transformer. In order to avoid ``attending" to the padding, masking is used. Masking is also used to enforce sequential dependencies, e.g.,  for our implementation, we enforce an L-R sequential dependence.
\end{enumerate}

We will elaborate on the architecture of our specific implementation later on in this paper. However, we present a high-level schematic of how multiple prediction tasks can be performed using a transformer in Fig. \ref{fig:transfomer}. Fig. \ref{fig:next} depicts a standard next-character prediction task from the literature. In this case, the outputs are softmax probabilities over the vocabulary. Fig. \ref{fig:cls} depicts using the transformer as a binary malicious/benign classifier -- the baseline ``decode-to-label" approach presented in Sec. \ref{sec:baseline}. Fig. \ref{fig:mixed} depicts the mixed objective approach presented in Sec. \ref{sec:mixed}, where a loss over the next character prediction output is used as an auxiliary loss in conjunction with the main classification task loss during training.

A downside of this architecture is that the transformer's multi-headed self-attention mechanism scales quadratically with respect to input sequence length. This is particularly problematic in the case of raw bytes of a binary file, whose dependencies can be spread over large swaths of the executable. Many different approaches to addressing these computational and memory inefficiencies have been proposed, ranging expanding the context window by ignoring tokens \cite{child2019generating,sukhbaatar2019adaptive}, reducing the representation in terms of memory and parameter size \cite{kitaev2020reformer,lan2019albert}, sequence-level recurrence \cite{dai2019transformer}, and even dropping attention altogether in favor of pre-trained convolutions \cite{tay2021pre}.

\subsection{Transformers for InfoSec}\label{transformer_infosec}

ML for InfoSec has been researched for decades \cite{rudd2016survey}, but widespread industry adoption has occurred only over the past several years. Of the adopted models, the majority of them utilize some form of hand-crafted features (e.g., \cite{rudd2018meade,rudd2019aloha,anderson2018ember,raff2019kilograms,kyadige2019learning,ducau2019smart}), with only a few operating on raw data. Sequence models, e.g., recurrent neural networks (RNNs) have been applied with some success \cite{pascanu2015malware}, but only to niche problems. Training RNNs for most problems is fundamentally not scalable to long sequences, both in terms of the hefty latency and memory requirements, %incremental training times 
and in terms of lack of the general loss of information from long term dependencies.

Such long term dependencies are common in NLP tasks, where it has become common to pre-train transformers in an auto-regressive manner on large unlabeled sequences of text, via next token prediction for L-R models and masked language modeling (MLM) for bidirectional models, then fine tune on smaller quantities of labeled data. This is often necessary, when potentially few labeled examples are available for the task of interest. However, many InfoSec applications where ML works well have millions to hundreds of millions of labeled samples. These labels are typically derived from an aggregation of multiple weak labeling sources \cite{ratner2017snorkel,rudd2018meade,fu2020fast}. 

We are not the first to apply transformers to InfoSec tasks. Li et al. applied transformers to malware detection in \cite{li2019mad}. However, contrary to their
hierarchical approach with multiple transformers and neural networks trained with different training/fine-tuning regimes,our approach is built on a single end-to-end transformer. Moreover, their approach also involves disassembly which adds intrinsic context for sequential modeling, but is expensive in terms of added processing time. Our approach by contrast, operates on raw bytes. While their work demonstrates superior cross-validated performance on assembly code representations,
their evaluation is severely limited in scale ($\sim$ 10k samples) and further evaluation is needed to assess its efficacy in realistic scenarios.

\newtext{Another approach by Pei et al. \cite{pei2020trex} uses a hierarchical transformer trained on dynamic micro-traces of a number of different functions in an MLM regime. The learnt representation is then fine-tuned to yield state-of-the-art results on a variety of semantic similarity tasks (e.g., determining semantic similarity of two functions across different architectures, compiler optimizations, etc.). The authors additionally provide a dataset for semantic similarity benchmarking. %While their approach is interesting, 
However, it differs from ours insofar as their hierarchical transformer uses micro-traces as its input modality whereas ours uses raw bytes.}

While transformers excel at sequence modeling, in this work, our predominant application is binary detection. In order to encourage the transformer to fuse sequential context with class information, in this work we combine the tasks of next character prediction and malicious/benign classification into a common loss function. Our approach is a natural extension of other research  \cite{rudd2019aloha,ducau2019automatic,huang2016mtnet,rudd2016moon} on multi-objective/multi-task training, which demonstrates that incorporating auxiliary losses can improve the performance of a classifier on the main task. In contrast to these previous approaches, however, our approach is applied to transformers and uses a novel technique for dynamically re-weighting the per-task loss, such that no auxiliary loss term ever dominates regardless of the loss magnitude.

We apply transformers to two InfoSec tasks: malicious URL detection and malicious PE file detection. 

Our work is not the first to address malicious URL detection -- there are several approaches in the literature that use convolutional architectures to classify malicious/benign URLs and domain generation algorithm ( DGA)-generated domains including \cite{sahoo2017malicious,saxe2017expose,le2018urlnet,yu2018character}. However, 
%While these works are interesting, note that 
they use different benchmark data sets, and are thus not directly comparable to our work. We benchmark against
similar, internally developed convolutional and non-convolutional architectures in this paper, 
%similar convolutional and non-convolutional architectures in this paper,  which we have developed internally. 
Note that the primary focus of this work is not creating the optimal URL classifier, but exploring how to train transformers for a typical non-NLP InfoSec machine learning task and contrasting with transformer training regimes for NLP problems.

With respect to detecting malicious PE files, several works have tackled this problem, e.g., \cite{rudd2019aloha,anderson2018ember,ducau2019automatic,kyadige2019learning,coull2019activation,raff2018malware}. While many apply classifiers fit over raw features (e.g., string hashes, byte entropy histograms, etc.), the most meaningful comparisons are models which operate on raw bytes, e.g., \cite{coull2019activation,raff2018malware}. However, there is an additional challenge when applying transformers to long sequences -- specifically, the size of the sequence context window that can fit in memory is substantially limited. There have been several methods proposed in the literature to ameliorate this, including modeling using sparse attention patterns partial span lengths, or utilizing approximations to avoid quadratic-complexity, 
%sparse attention patterns and partial span lengths and utilizing approximations to avoid full attention computations, 
e.g., \cite{lu2019vilbert,sukhbaatar2019adaptive,beltagy2020longformer,kitaev2020reformer,dai2019transformer}. We apply our own variation of the adaptive span method proposed by Sukhbaatar et al. \cite{sukhbaatar2019adaptive} in Section \ref{sec:byte_transformer}.

% \begin{figure}[!t]
%     \centering
%     \includegraphics[scale=0.45]{images/multihead.png}
%     \caption{Multi-head Attention Operation.}
%     \label{fig:multihead}
%      \vspace{-0.5cm}
% \end{figure}

%% file: 3_Method.tex
\section{URL Transformer Approach}
\label{sec:approach}

Let $X$ be a dataset of URLs with binary labels $Y$. Let $x$ be a generic URL from $X$ with label $y \in \{0,1\}$, 0 corresponding to benign and 1 corresponding to malicious. 

\subsection{Baseline: Decode-to-Label}
\label{sec:baseline}

We refer to our baseline transformer approach as ``decode-to-label", where the transformer ingests a URL, and only the hidden state corresponding to the a special classification token ($\langle$CLS$\rangle$) is used to predict the label. Contrary to sequence transduction tasks that leverage an encoder transformer and a decoder transformer to encode a sequence into another sequence (e.g., language translation), we can think of this approach as using a transformer strictly as a decoder from a source sequence to a malicious/benign label. Our ``decode-to-label" approach uses a left-to-right (L-R) decoder, similar to OpenAI's GPT \cite{radford2018improving} approaches, with a $\langle$CLS$\rangle$ token placed at the end of the sequence. Sequence information is propagated through the attention layer states in a L-R manner, meaning that hidden state $k$ in the $l$th attention layer is fed information corresponding to hidden states $1,2,\hdots,k-1,k$ in the $(l-1)$th layer. The final classification is made via a feed-forward neural network (FFNN), which is fed the top layer's hidden state corresponding to the $\langle$CLS$\rangle$ token. The final dense layer of the FFNN projects the output to a 1D value. This value is then passed through a Sigmoid activation function to assume a final prediction  $h(x) \in [0,1]$. Binary cross entropy between the prediction $h(x)$ and label $y$ is evaluated during training and the associated gradients are backpropagated. The associated \textit{Classification Loss} is then:

\begin{equation}
    L_{CLS}(x,y) = -y log(h(x)) + (1-y)log(1-h(x)).
\end{equation}

In contradistinction to most transformer literature, this approach does not utilize any explicit loss over hidden states corresponding to tokens within the sequence. Our rationale for applying the ``decode-to-label" approach is to provide a benchmark against which to assess any gains and losses that have been introduced by explicitly optimizing sequential information into the transformer. 

\subsection{Next Character Prediction Pre-Training and Fine-Tuning}
\label{sec:next}

Next character prediction tasks have been applied throughout the transformer literature. For this regime, no labels are used during pre-training. Instead, we have a pre-training set $\hat{X}$ which may or may not have associated malicious/benign labels. Each subsequent character in the URL serves to label each previous character. For example, given input URL

\begin{center}
$\hat{x}=$\url{https://www.xyzsecurity.com/},
\end{center}

\noindent
 the ``label" sequence would be

\begin{center} 
\url{ttps://www.xyzsecurity.com/}$\langle$CLS$\rangle$. 
\end{center}

%\begin{center}
%$\hat{x}=$\url{https://www.fireeye.com/},
%\end{center}

%\noindent
% the ``label" sequence would be 

%\begin{center} 
%\url{ttps://www.fireeye.com/}$\langle$CLS$\rangle$. 
%\end{center}
 
Note that the $\langle$CLS$\rangle$ token is omitted from the input sequence. 

We encode each character with its respective ASCII byte value ranging from 0 to 255. Note that in practice only a subset of these byte values are manifest in our data. We use the value 256 to represent our $\langle$CLS$\rangle$ token, yielding 257 distinct input embeddings.

Next character prediction is performed over each hidden state of the transformer output, corresponding to the embedded vector of the sequence up until that point. A feed-forward neural network (FFNN), which takes the corresponding transformer hidden state as input is used as a predictor of the next character. The output of the FFNN is a 257-element softmax, with the first 256 output probabilities corresponding to the probabilities of specific byte values as the next character and the last output probability corresponding to the probability of the $\langle$CLS$\rangle$ token (i.e., the end of the sequence). Note that for the next character prediction task, the input sequence $\hat{x}$ ignores the $\langle$CLS$\rangle$ token at input; this token is only used as a ``label" for the last character of the URL.

Loss is evaluated as the categorical cross entropy over the entire sequence, normalized by the sequence length $M$. Let $I(\cdot)$ be an indicator function which evaluates to 1 if the argument is true and 0 otherwise. The next character loss function over the URL $\hat{x}$ becomes:

\begin{equation}
L_{NEXT}(\hat{x}) = -\frac{1}{M}\sum_{i=1}^{M}\sum_{j=0}^{256} I(\hat{x}_{i+1} = j) log (h(\hat{x}_i)_j).
\end{equation}

Backpropagation of $L_{NEXT}$ is used to train the underlying transformer representation. During fine-tuning, the pre-trained representation is loaded, potentially with lower attention layers frozen. With a decreased learning rate, $X$ and $Y$ are fed to the transformer with $L_{CLS}$ used to train a malicious/benign predictor and update the unfrozen parameters of the transformer.  

\subsection{Balanced Mixed Objective Training}
\label{sec:mixed}

This approach aims to jointly optimize for both next character prediction and malicious/benign classification across dataset $X$ with labels $Y$. Note that this does not preclude next character prediction pre-training over another dataset. The rationale behind this approach is built on prior research which suggests that optimizing over multiple (correlated) tasks simultaneously leads to a better performing classifier with more stable convergence characteristics \cite{rudd2019aloha}.

Following this rationale, we apply a mixed objective optimization approach, which balances main malicious/benign determination task with an auxiliary next character prediction loss. Contrary to previous research, which uses ad-hoc fixed weights on main and auxiliary task losses, our novel approach employs an adaptive balancing scheme, which ensures that no single loss term dominates, regardless of the loss value. Our loss-weighting strategy is as follows: 
 
\begin{align}
\label{eq:mixed}
\begin{split}
    % L_{MIXED} = &\alpha(a,L_{CLS},L_{NEXT}) L_{CLS} \\
    %             + &\beta(b,L_{CLS},L_{NEXT}) L_{NEXT},
    L_{MIXED_I} = &\alpha_I L_{CLS_I}+ \beta_I L_{NEXT_I}.
\end{split}
\end{align}
\noindent

At iteration $I$ of training, corresponding to one mini-batch, values $\alpha_I$ and $\beta_I$ are balancing multipliers computed for each mini-batch, and are assumed constant when computing the gradient of the loss function. They are used to weight the classification and next character prediction loss components.
Respectively, $\alpha_I$ ensures that $L_{next_I}$ accounts for $\frac{a}{a+b}$ of $L_{mixed_I}$ and $\beta_I$ ensures that $L_{cls_I}$ accounts for $\frac{b}{a+b}$ of $L_{mixed_I}$. Values $a$ and $b$ are hyperparameters which we fix during training.
%$\alpha_I$ and $\beta_I$ are balancing multipliers computed to weight the classification and next character prediction loss components. 
Note that for simplification we can normalize such that $a+b:=1$ and say that $a$ and $b$ are the respective loss fractions themselves.  For our experiments, we set $a:=b:=0.5$ unless specified otherwise. Values $\alpha_I$ and $\beta_I$ are computed for each minibatch, according to:

% Thus, $\alpha_I$ and $\beta_I$ are balancing multipliers which ensure that $L_{next}$ accounts for respectively $\frac{a}{a+b}$ and $\frac{b}{a+b}$ of the net loss contribution to $L_{MIXED}$. Values $a$ and $b$ are hyperparameters. For our experiments, we set $a=b=0.5$ unless specified otherwise. Values $\alpha$ and $\beta$ are computed for each minibatch, according to: 

\begin{align}
    \alpha_I &= \frac{a(L_{CLS_I}+L_{NEXT_I})}{L_{CLS_I}},\\
    \beta_I &= \frac{b(L_{CLS_I}+L_{NEXT_I})}{L_{NEXT_I}}.
\end{align}

Note that in this work, we apply only two loss types. However, we could trivially extend our approach to $K$ different loss types as follows. Given generic multiplier $\gamma_{iI}$ for the $i$th loss term at iteration $I$, $L_{iI}$, and desired loss contribution fraction $c_i/\sum_{j=1}^K c_j$, we compute $\gamma_{iI}$ as follows: 

\begin{equation}
\gamma_{iI} = \frac{c_i\sum_{j=1}^K L_{jI}}{L_{iI}}.
\end{equation}

%% file: 3.1_Byte_Approach.tex
\section{Byte Transformer Approach}
\label{sec:byte_transformer}

As generic byte sequences tend to be much longer than URLs, the stock transformer architecture utilized in Sec. \ref{sec:approach} quickly becomes infeasible. There are a multitude of approaches for extending context window sequence lengths in the literature (see Sec. \ref{sec:background}). Of these approaches, the notion of adaptive attention span, introduced in \cite{sukhbaatar2019adaptive}, is intriguing. Moreover, the implementation by \cite{sukhbaatar2019adaptive} is relatively mature and less stringently designed around specific NLP tasks. Thus, we chose this method as the basis for a feasibility study of a Byte Transformer. Note, however, that any of the plethora of other techniques discussed in Section \ref{sec:background} could be  employed to this end.

Adaptive span transformers conserve memory by applying the attention operation only over a prior sub-span of the context window, similar to Sparse Transformers. Theoretically, over multiple layers of the Attention + FFNN stack, information from the entire sequence should percolate into the final attention layer, despite the reduction in context window size. The Adaptive Attention Span paper also incorporated a parameter that tunes the respective attention span of each layer in the Attention + FFNN stack. 

\begin{figure}[!t]
  \centering
      \includegraphics[width=\linewidth]{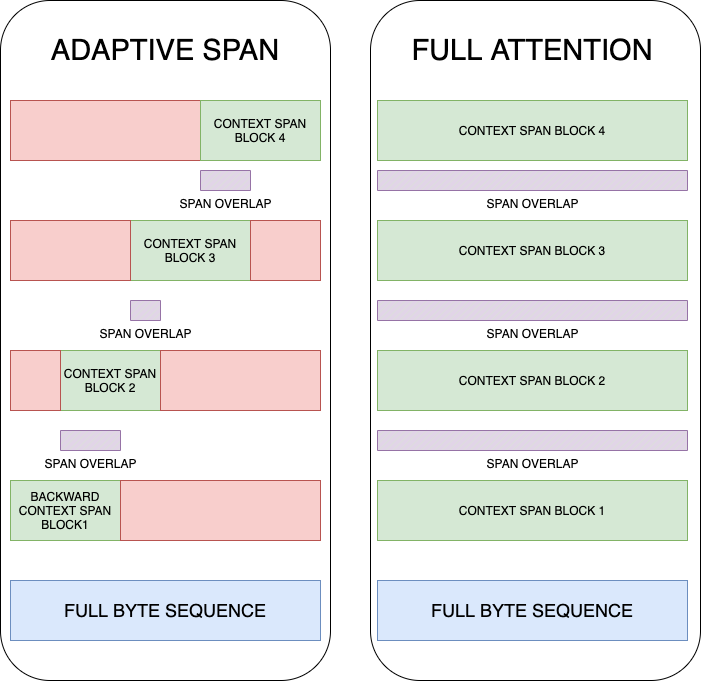}
  \caption{A simplified schematic depicting intuitively how the Adaptive Attention Span approach leads to memory savings over conventional "Full Attention". Memory savings are depicted in red, attention spans are depicted in green, and context overlaps are depicted in purple. See text for additional information.\label{fig:Intuition}}
\end{figure}

A simplified schematic, aimed to capture the intuition behind Adaptive Span Transformers is presented in Fig. \ref{fig:Intuition}. Here, four sampled (not necessarily subsequent) blocks of Attention + FFNN are depicted in two separate stacks; one from an Adaptive Span transformer and one from a canonical ``Full Attention" transformer. For the Adaptive Span transformer, the overlap in span from hidden states between prior and subsequent blocks notionally propagates sequence information over the full context of the input sequence as Attention + FFNN blocks are stacked. If partial span information can be successfully fused over the full stack of Attention + FFNN blocks, then equivalent performance to a canonical ``Full Attention" transformer can be obtained at significantly reduced memory requirements. In \cite{sukhbaatar2019adaptive}, Sukhbaatar et al. validate this approach for canonical NLP problems, and we employ their approach as well as a spinoff of their implementation for our Byte Transformer.

We modified the original code from Sukhbaatar et al. by removing the log softmax multinomial output (for next character prediction tasks) and placing a dense layer with a sigmoid output atop the final hidden state. While we kept the hidden state caching functionality from Dai et al., which allows for tackling even longer sequences \cite{dai2019transformer}, we did not use this functionality during our experiments on EMBER (see Sec. \ref{sec:ember}), as this 1) makes training on batches of data more difficult for varying length sequences and 2) takes longer to train. We note, however, that on our infrastructure, combining state caching with reduced attention spans (see Sec. \ref{sec:ember}), we were able to feasibly train on sequences of up to 20kB.

During training, we used generators which enforced balancing of malware and goodware samples within each minibatch, with a minibatch size of 32. For our optimizer, we used SGD, with a learning rate of 0.005, momentum of 0.9, and a decay rate of 0.0001. \newtext{For our Byte Transformer experiments, we utilized only the classification loss without jointly optimizing next-token prediction (see~Fig.~\ref{fig:transfomer}).}

%% file: 4_Evaluation.tex
\section{URL Experiments}

We collected a dataset of URLs, down-sampled over 2-3 months from a threat intel feed in late 2019. Each URL had multiple weak malicious/benign labels. We derived a single malicious/benign label for each URL using Snorkel \cite{ratner2017snorkel}. The training set consisted of 1,007,451 labeled URLs with 180,052 malicious and 826,333 benign. The validation set consisted of 111,930 URLs with 20,013 malicious and 91,813 benign. The test set consisted of 279,874 URLs with 50,029 malicious and 229,604 benign respectively. We additionally collected a dataset of 20 million unlabeled URLs for pre-training experiments, disjoint from train, test, and validation sets. Note that we did not perform any cleanup of these URLs, meaning that our dataset includes lengthy/obscure URLs as well as raw IPs; thus the reported performance numbers are not comparable with those from other authors with proprietary datasets (e.g., \cite{saxe2017expose}).

\subsection{Transformer Base Topology}

We implemented a transformer with 20 layers of attention, a context window size of 256, a model hidden state size of 64, a feed-forward dimension of 128, and 4 attention heads per-layer. We also employed dropout, with a dropout ratio of 0.1. For classification, we apply a feed forward neural network (FFNN) atop the output corresponding to the $\langle$CLS$\rangle$ token. This FFNN consists of a 64-dimensional input followed by Layer Normalization, a hidden layer of 32-dimensions, Exponential Linear Unit activation, another hidden layer, reducing dimensionality from 32 to 1, and a sigmoid output. For next character prediction, we apply a similar architecture, but with a softmax output of 257 dimensions, predicting bytes 0 through 255 and the $\langle$CLS$\rangle$ token (which we ascribe label 256).

\subsection{Comparison of Different Training Regimes}

We performed comparisons of our four training techniques, ensuring that performance converged on the validation set.  

\paragraph{Baseline (DecodeToLabel):}
%{\bf Baseline (DecodeToLabel):} 
As a baseline, we compute loss only over the binary malicious/benign prediction, performing neither pre-training nor next character prediction. We trained the model for 15 epochs with minibatch size 512.  For this training regime we used PyTorch's default Adam optimizer.

\paragraph{Auto-regressive Pre-training and Fine Tuning -- Training Set (FineTune):}
%{\bf Auto-regressive Pre-training and Fine Tuning -- Training Set (FineTune)} 
For this experiment, we first performed pre-training over the entire training set for the next-character prediction task for 15 epochs. We then performed fine-tuning, freezing the first 16 attention layers and trained the decode-to-label task for 15 epochs at a reduced learning rate. For Pre-Training, we used PyTorch's default Adam optimizer. For fine-tuning, we used SGD \newtext{optimizer} with a learning rate of 1e-4.

\paragraph{Auto-regressive Pre-training and Fine Tuning -- Pre-Training Set (FineTune 20M):}

%{\bf Auto-regressive Pre-training and Fine Tuning -- Pre-Training Set (FineTune 20M)} 
\oldtext{For this experiment, we first performed pre-training over the 20 million URL pre-training dataset for 2 epochs, using PyTorch's default Adam optimizer.}\newtext{For this experiment, we first performed pre-training over the 20 million URL pre-training dataset using PyTorch’s default Adam optimizer. Only 2 epochs were required for convergence using a dataset of this magnitude.} We then performed fine-tuning, freezing the first 16 attention layers and trained the decode-to-label task for 15 epochs at a reduced learning rate. For fine-tuning, we used SGD \newtext{optimizer} with a learning rate of 1e-4.

\paragraph{Balanced Mixed Objective Training (MixedObjective):} 
We performed 15 epochs of balanced  mixed objective training over the training set using PyTorch's default Adam optimizer.

\subsection{URL Transformer Results}

We report results in terms of Receiver Operating Characteristics (ROC) curves and area under the ROC curve (AUC). 

\begin{figure}[!t]
  \centering
      \includegraphics[width=\linewidth]{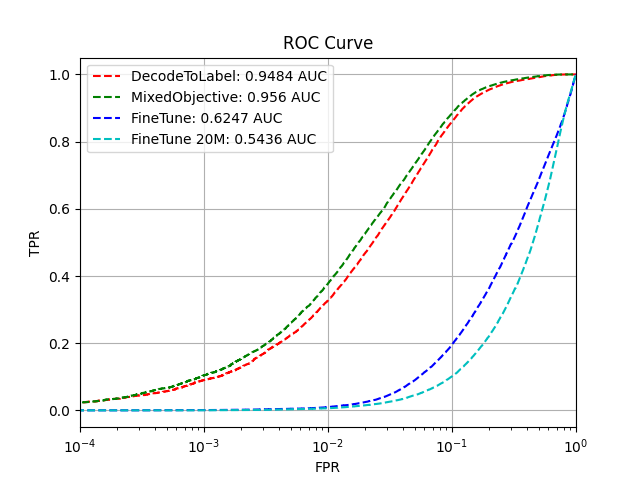}
  \caption{ROC and ROC-AUC results from different URL transformer training approaches. The Mixed Objective approach, shown in green, outperforms a generic decode-to-label approach. Neither fine-tuning approaches yielded impressive performance.\label{fig:PerformanceComparison}}
\end{figure}

Surprisingly, fine-tuning from a pre-trained initialization resulted in poor performance compared to training from scratch from a Xavier initialization (see Fig. \ref{fig:PerformanceComparison}). We additionally attempted fine-tuning, freezing all but the last 4 layers of the transformer, but witnessed similar performance to our original fine-tuning regime.  This is despite a consistent decrease and general convergence in loss for both of the pre-trained representations. This suggests that there is less immediate task transfer between next-character prediction for URLs than there is for next-character prediction in NLP contexts. 

Despite the failure of pre-training with an auto-regressive loss to deliver performance gains, we did find that we were able to achieve marginal performance improvements by introducing an auxiliary next-character prediction loss. This is consistent with the findings of Rudd et al. \cite{rudd2019aloha}.

\newtext{While both $L_{NEXT_I}$ and $L_{CLS_I}$ from Sec.~\ref{sec:approach}  tended to converge with training, we noticed $L_{NEXT_I}$ was consistently an order of magnitude greater than $L_{CLS_I}$. This is not surprising, as the classification output predicts a single binary malicious/benign score from the $\langle$CLS$\rangle$ token embedding, while the next character prediction loss is summed over the full length of the URL. While the re-scaling in the joint loss term in Eq. \ref{eq:mixed} prevents either $L_{NEXT_I}$ or $L_{CLS_I}$ from having an undue influence over the optimization process, this may affect the rate at which $L_{NEXT_I}$ and $L_{CLS_I}$ converge but does not directly affect the magnitude of $L_{NEXT_I}$ or $L_{CLS_I}$.}

\subsection{Comparison Models}

As a viability comparison, we benchmarked our transformer against several other models which we have developed for malicious/benign URL detection. Substantial development and testing effort went into these models. We summarize them in this section.

\paragraph{Random Forest on SME-Derived Features:} For this model, proprietary features were derived with the help of subject matter experts (SMEs). These feature vectors consist of binary values/counts derived from parsing the URL and checking if specific parsed values from the URL string reside in various lists corresponding to likely indicators of malicious or benign content. The derived feature vectors therefore characterize the content of the URL. 

We then fit a random forest on these extracted feature vectors. The random forest classifier consisted of 30 trees, each with a maximum depth of 20. Nodes were split based on information gain. For all other parameters, we used Scikit-learn's \cite{pedregosa2011scikit} defaults.

\paragraph{LSTM on Raw URLs:} For this model, we fit a long short-term memory neural network (LSTM) over the URLs, using embeddings of size 50 and an LSTM hidden size of 100. We performed optimization for a max of 100 epochs with early stopping based on validation performance. During optimization, we used an Adam optimizer with Keras's default parameters and a minibatch size of 128.

\paragraph{1D CNN on Raw URLs:} This model first embeds each input token into a 40-dimensional vector, then follows up with a dropout layer (p=0.2) and a stack of two 1D convolutions with ReLU activations. The convolutional layers have 256 and 100 filters respectively with kernel sizes 5 and 3. The output of the convolutional layer stack is then globally max-pooled and passed through a hidden size layer of size 256, a dropout layer (p=0.2), and a ReLU activation. This is then transformed to 1D via a final fully-connected layer and passed through a Sigmoid output.

% {\bf 2D CNN on Raw URLs} This model first embeds each input token into a 20-dimensional space and then expands the input to a rank-3 tensor. Two blocks of 32 3x3 2D convolutional filters followed by ReLU activations are then stacked, the output of the stack is max-pooled, and then dropout (p=0.25) is applied followed by a dense layer compressing to 512-dimensions, ReLU activation, and then a final fully connected layer and sigmoid activation. The same optimization strategy used in the LSTM is applied.

\begin{figure}[!t]
  \centering
      \includegraphics[width=\linewidth]{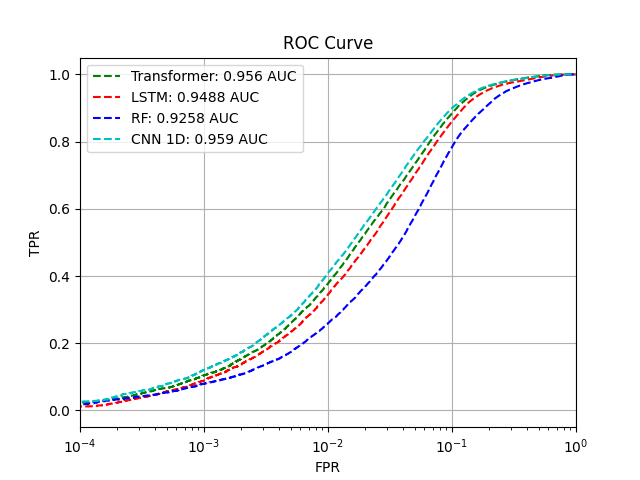}
  \caption{ROC and ROC-AUC results for comparative baselines along with our top performing transformer model. Note that the Mixed Objective transformer approach performs on-par with the highest performing baseline (CNN 1D), outperforming all other approaches.\label{fig:URLBaselines}}
\end{figure}

Our mixed objective transformer model (MixedObjective in Fig. \ref{fig:URLBaselines}) outperforms all but one of the comparison models, substantially outperforming the random forest feature-based model (RF) and the LSTM and performing on-par with the CNN model. 

\newtext{As previously stated, our URL results are not directly comparable to other works, as they are derived from a different dataset. Our URL dataset was collected with the use case of checking embedded URLs in  suspicious emails in mind, and using the results as indicators to flag an email as suspicious or warn around a specific link contained within. The URLs chosen were unfiltered and intentionally representative of more challenging edge cases in a production pipeline, where ML detection is one of many tools in the defensive arsenal. Thus, the ROC curves presented herein should not be read as performance of a standalone commercial system. With respect to URL detection rates, this is a more general and more difficult use-case than checking URLs strictly meant to appear as human-readable content, e.g., a spoofed payment processing page. This is also a different and potentially more difficult problem than simply detecting URLs/domains generated by domains generation algorithms (DGAs), as the content on a website is incidental to how the domain was generated, i.e., DGA-generated domains are not necessarily malicious and non-DGA-generated domains are not necessarily benign. However, our results are consistent with other literature in terms of rank order of similar classifiers. Saxe and Berlin \cite{saxe2017expose} found that a 1D CNN on character embeddings (similar to ours) outperformed a baseline trained on extracted features, while Yu et al. \cite{yu2018character} found that this same architecture yielded state of the art for DGA detection, slightly outperforming LSTM models and significantly outperforming random forests over lexical features.}

% We will include a statistical significance test in the camera-ready version, performing multiple stochastic initializations if this paper is accepted, but we suspect that the top-performing CNN (CNN 1D in Fig. \ref{fig:URLBaselines}) and transformer model performances are statistically identical.

%% file: 4.1_Byte_Evaluation.tex
\section{PE Experiments}

In Sec. \ref{sec:ember} we examine the viability of our Byte Transformer via experiments on the EMBER 2018 PE dataset \cite{anderson2018ember}. We then investigate encoding schemes to potentially improve the effectiveness of byte-based classifiers in Sec. \ref{sec:byte_pair} and Sec. \ref{sec:kilograms}.

\subsection{Performance Comparison on EMBER}
\label{sec:ember}

Due to GPU memory imposed constraints on transformer context windows, even for adaptive span transformers, we truncated each of the EMBER PE files to 4096 bytes. Thus, performance comparisons should be read in terms of relative rank of each benchmark, rather than compared to  numbers in the original EMBER dataset. \newtext{This was the only change that we made to the EMBER benchmark; for consistency, we maintained the original temporal dataset splits and followed the training and evaluation protocol from \cite{anderson2018ember}.} We conducted training using a Cirrascale server with 8 Tesla M40 24GB GPUs. 

Note that we were able to effectively expand the context window size to 20kB using the TransformerXL caching mechanism \cite{dai2019transformer}, but opted not to employ this in our experiments as it significantly increased training time, and added additional levels of complexity. 

We trained our Byte Transformer on the EMBER training set using the following selection of hyperparameters. Note that this was one of a few hyperparameter choices that actually attained convergence:

\begin{itemize}
    \item Hidden Size: 32
    \item Inner Hidden Size: 128
    \item \# Layers: 10
    \item Block Size: 256
    \item \# Heads: 2
    \item Attention Span: 2048
    \item Dropout: 0.05
    \item Embedding Dropout: 0.05
\end{itemize}

For the classifier, we used a logistic regressor fit atop the 32-dimensional hidden state corresponding to a $\langle$CLS$\rangle$ token. Logistic regression weights and transformer parameters were jointly optimized during training. We trained the Byte Transformer for 15 epochs \newtext{after which validation convergence was attained}.

To assess the relative performance of our Byte Transformer model, we fit three benchmark classifiers. Two of these classifiers were variations of the MalConv and LightGBM models from the original EMBER paper, with minor modifications for a fair comparison. For MalConv, we limited the context to 4096 bytes, and for the GBM model, we modified the EMBER feature extractors to only extract from the first 4096 bytes of each file. 

Note that 4096 bytes is the approximate size of a typical PE header. \newtext{In practical adversarial settings, an attacker could bypass this model by ensuring that malicious byte sequences manifest beyond the extent of the file’s initial 4096 bytes. However, real-world industry malware detection approaches commonly employ defense-in-depth strategies in which this model would be one of many complementary detection methods to help counteract such an evasion. Nevertheless, future work towards extending this truncation window using other efficient Transformer architectures \cite{dai2019transformer,child2019generating,kitaev2020reformer} represents a promising research path forward from this feasibility study.} 

We additionally employed a deep convolutional neural network that combines a 10-dimensional, learnable embedding layer with a series of five interleaved convolutional and max-pooling layers arranged hierarchically so that the original input size is reduced by one quarter (1/4) after each layer \cite{coull2019activation}. We refer to this as the MG-CNN.

\begin{figure}[!t]
  \centering
  \includegraphics[width=\linewidth]{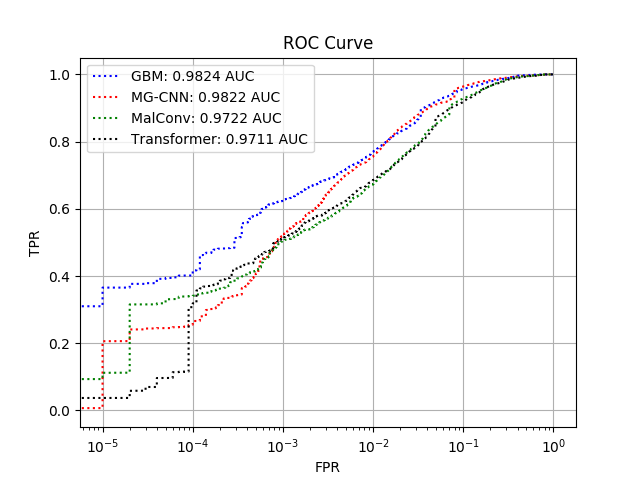}
  \caption{Performance of different classifiers on the byte-truncated version of the EMBER dataset.\label{fig:EMBERComparisonRaw}}
  \vspace{-0.5cm}
\end{figure}

 In other papers, byte-based models often truncate inputs beyond a maximum length, while feature-based models use features extracted from the entire file irrespective of length, potentially presenting an overly optimistic view of feature-based classifiers. Our comparison, by contrast, is fair in the sense that inputs to each respective model are extracted using the same context window length. Results of this comparison are shown in Fig. \ref{fig:EMBERComparisonRaw}. We see that our Byte Transformer and MalConv models perform comparably, but under-perform the MG-CNN and lightGBM models, which also perform comparably. This suggests that there is a  motivation for the extra layers added in the MG-CNN over MalConv, and that much of the information learned during training may overlap with content extracted from EMBER features. This finding is consistent with \cite{demetrio2019explaining}, where the authors found that a convolutional model trained on EMBER largely uses portions of the header when discriminating betweeen malicious/benign.
 
\subsection{Effects of Byte Encoding Schemes}
\label{sec:byte_pair}

Note that both our Byte Transformer and the MG-CNN share two common characteristics: First, their convergence is extremely sensitive to hyperparameter selection. Second, their practical application is somewhat hindered by limited context window sizes imposed by memory constraints (though the MG-CNN can feasibly incorporate far longer sequences). Since successful applications of transformers on raw data in the NLP space leverage specific encoding strategies, e.g., RoBERTa models leverage a byte-pair encoding \cite{liu2019roberta}, we explore whether similar encoding schemes can potentially benefit InfoSec byte-based models. 

To shorten training times and perform this investigation using more realistic context windows, we performed the following comparisons on the full EMBER dataset using the MG-CNN model with each sample truncated at 102400 characters and different encoding algorithms:

\begin{itemize}

\item {\bf Baseline (Baseline):} MG-CNN model fit on raw bytes.

\item {\bf Byte Pair Encoding (Byte Pair):} For this, we performed 10 iterations, greedily adding 10 characters per iteration. This lead to a total vocab size of 396. We first truncated our input at 102400 bytes and then performed the byte pair encoding, which compressed each input at the expense of increased vocab size.

\item {\bf Removal of Common Padding Sequences (No Pad):} We removed padding bytes for 0xFF and 0x00 for sequences longer than 3 subsequent padding bytes of the same type and replaced them with a sequence of length 3. As with the byte pair encoding, we first truncated at 100kB and then performed the compression.

\item {\bf Byte Pair Encoding Beyond 100 kB (Byte Pair Extra):} Same encoding as byte pair, but performed encoding on the full file then truncated to 102400 tokens. This has the practical effect of extending sequence length.

\item {\bf Pad Removal Beyond 100kB (No Pad Extra):} Removed padding from the entire file first then truncated to 100kB.
\end{itemize}

Surprisingly, none of our compression schemes improved performance beyond the baseline. This is specifically surprising, as the {\bf byte pair extra} model introduces no information loss, yet incorporates longer contexts from the original sequence.

\begin{figure}[!t]
  \centering
      \includegraphics[width=\linewidth]{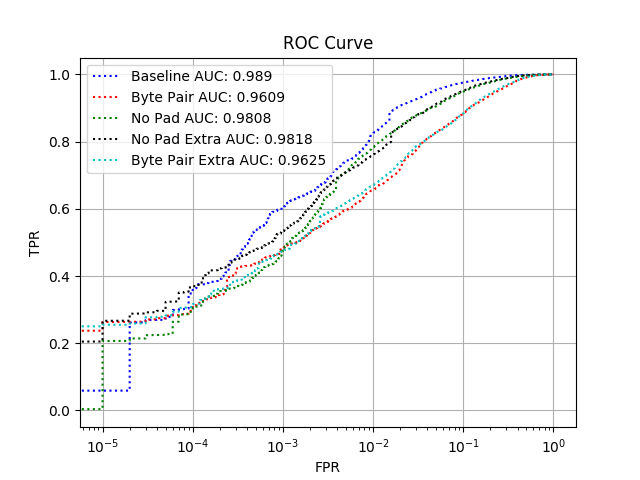}
  \caption{Performance of the MG-CNN classifier on the EMBER dataset under various encoding/compression regimes.\label{fig:EMBERByteCompression}}
\end{figure}

\subsection{Effects of Kilo-Gram Encoding Schemes}
\label{sec:kilograms}

In this section, we examine to what extent we can effectively extend the length of a context window by either removing common kilo-gram~ \cite{raff2019kilograms} sequences and keeping all other bytes, i.e., blacklisting kilograms, or by keeping only common kilo-gram sequences and removing all other bytes, i.e., whitelisting kilo-grams.

For these experiments, we first flagged kilograms for each sample up to 12288 bytes, performed whitelisting and blacklisting by keeping only or removing only these kilograms respectively, then truncated everything in each sequence beyond 4096 bytes. We chose 6-grams because these seemed to work well both in the kilograms paper and as features for some of our unpublished detection models.% some of our internal models.

\begin{figure}[!t]
  \centering
  \includegraphics[width=\linewidth]{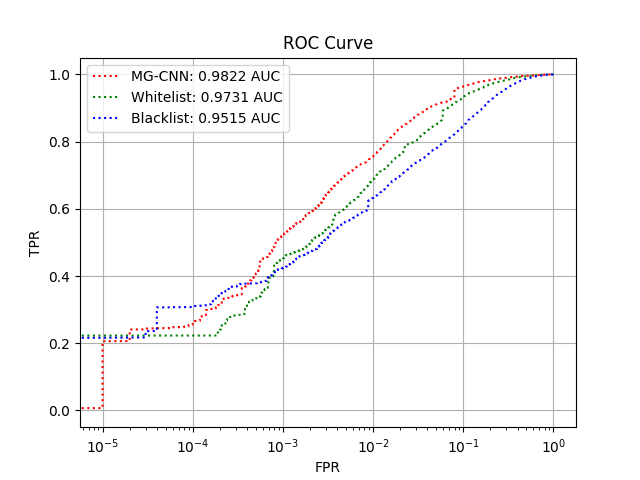}
  \caption{Performance of different classifiers on the byte-truncated version of the EMBER dataset blacklisting/whitelisting kilograms.\label{fig:EMBERKilogramsCompression}}
  \vspace{-0.5cm}
\end{figure}

Performance comparisons of baseline, whitelist, and blacklist models are shown in the Fig. \ref{fig:EMBERKilogramsCompression}. Each approach utilized an MG-CNN model.

Neither blacklisting nor whitelisting improved classification performance, despite the fact that broader byte information from larger context window spans get packed into the input for the MG-CNN. Since whitelisting yielded a better classifier than blacklisting, this potentially suggests that 1) broader context delivered by kilogram whitelisting is important and 2) kilo-grams associated bytes  have greater signal content than non-kilogram associated bytes. This is consistent with the finding that kilograms serve as a useful feature representation for a variety of InfoSec problems.

%% file: 5_Discussion.tex
\section{Discussion}

We have demonstrated that transformers can achieve performance comparable to or better than that of other top-performing models for URL classification. We have also found that, contrary to NLP domains, wherein auto-regressive pre-training substantially enhances performance in a fine-tuned regime, for our URL data, auto-regressive pre-training on a large corpus yields no apparent gains for the classification task and makes it substantially more difficult to fit a performant model. This suggests that the next character prediction task has too little apparent correlation with the task of malicious/benign prediction for effective/stable transfer. Interestingly, utilizing next character prediction as an auxiliary loss function \cite{rudd2019aloha} yields improvements over training solely to predict the label. Note that this occurs even with a relatively large portion of the overall loss term (50 \%) devoted to the auxiliary loss. This suggests that mixed objective optimization is more effective at correlating across heterogeneous loss terms than fine-tuning.% Moreover, when tuning our model, consistent with findings by Rudd et al. in \cite{rudd2019aloha}, we found that auxiliary loss optimization \textit{improved} convergence speed and stability.

\oldtext{
Note that in contrast to some of the comparison models we did not perform a rigorous hyperaparameter search for our transformer, since this research was primarily concerned with loss functions and training regimes; not an ``optimal" model topology. We leave this to future work, but the relatively impressive performance obtained herein suggests that  transformers could likely achieve state-of-the-art on the URL prediction.
}

\newtext{Note that in this work we did not perform rigorous hyperparameter searches for either the URL transformer or the byte transformer. Given a motivation to use transformers for these or other problems, frameworks like Optuna~\cite{optuna} could be utilized to arrive at a better selection of hyperparameters which could very likely improve performance for both tasks studied herein. However, such a hyperparameter search would require considerable compute time/resources, and would be, in many respects, beyond the scope of this paper. This paper aims to explore the viability of transformers for end-to-end infosec tasks, not arrive at the some “optimal” transformer topology. This is one of the reasons why our architectural choices were strongly guided by well-studied approaches in the literature. We would also note that there are myriad other architectures in the literature that could be applied in our experimental settings.}

%% file: 6_Conclusion.tex
\section{Conclusion}

We have demonstrated first steps for training transformers for InfoSec tasks from scratch and applying them to two heterogeneous InfoSec datasets: malicious/benign URLs and goodware/malware PE binaries. 

While these datasets are not representative of all data in the ML for InfoSec space, they do reflect the common property of having a multitude of samples with labels derived from weak labeling sources, differing from common NLP classification tasks. 

For URLs, we have shown that unlike in NLP domains, there is little benefit to pre-training, but substantial benefit from mixed objective optimization. To this end, we have introduced a novel loss function which dynamically re-balances gradients of auxiliary losses with the main task loss at each training step, potentially improving training stability. Utilizing this loss function, our URL transformer model performs on-par with our top-scoring benchmark model. Performance could potentially be further improved using a bi-directional information flow trained under an MLM-like regime. 

For PEs, we have shown that we can successfully train adaptive span transformers from scratch on raw byte sequences, and for a limited context length perform on par with a Malconv model. Certain methods for extending this context length are trivial, e.g., via caching hidden states, and incorporating other scaling tricks could be feasible with some alterations (e.g., via a longformer-like locality-sensitive hashing approximation~\cite{beltagy2020longformer}), though these may lengthen training times and introduce additional complexity. 

While we have laid out first principles for training transformers for InfoSec tasks, there is still much work required to make these models widely viable: First, transformers are extremely compute-intensive compared to other models that operate on raw data. For the URL problem, a transformer would need to substantially outperform top-performing convolutional models in order to justify extra compute costs during training and deployment. While this is also the case for PE binaries or other byte sequence tasks, for these tasks, context length is also a problem that is difficult to fully address even with special tricks (sparse alterations to attention, adaptive span, hidden state caching, etc.).

We are presently working to extend our models to other data formats, including ways to incorporate even longer context and different training regimes that may improve classification performance.

%% file: 7_Acknowledgement.tex
\section*{Acknowledgments}
This research was funded by Mandiant Inc. Mandiant is a world-renowned cybersecurity company that specializes in cyber threat intelligence, cybersecurity data science, and cyber analytics.